%% file: main.tex
\PassOptionsToPackage{table}{xcolor}
\documentclass[10pt,twocolumn,letterpaper]{article}
\usepackage[pagenumbers]{iccv} %

\input{preamble}

\definecolor{iccvblue}{rgb}{0.21,0.49,0.74}
\usepackage[pagebackref,breaklinks,colorlinks,allcolors=iccvblue]{hyperref}

\title{Splat-LOAM: Gaussian Splatting LiDAR Odometry and Mapping}

\author{
Emanuele~Giacomini\textsuperscript{1}\qquad
Luca~Di~Giammarino\textsuperscript{1}\qquad
Lorenzo~De~Rebotti\textsuperscript{1}
\vspace{0.3cm}
\\
Giorgio~Grisetti\textsuperscript{1}\qquad
Martin~R.~Oswald\textsuperscript{2}
\vspace{8px}\\
{
\normalsize
\textsuperscript{1}{Sapienza University of Rome} \qquad
\textsuperscript{2}{University of Amsterdam}
}
}

\begin{document}
\maketitle
\input{sec/0_abstract}

\input{sec/1_intro}

\input{sec/2_related_work}
\input{sec/3_method}

\input{sec/4_experiments}
\input{sec/5_conclusions}
{
    \small
    \balance
    \bibliographystyle{ieeenat_fullname}
    \bibliography{main}
}
\input{sec/6_supp}

\end{document}

%% file: preamble.tex
\usepackage{amsmath}
\usepackage{bbm}
\usepackage{acro}
\usepackage{bm}
\usepackage{booktabs}
\usepackage{adjustbox}
\usepackage{multirow}
\usepackage{float}
\usepackage{transparent}
\usepackage{multicol}
\usepackage{balance}

\usepackage{glossaries}
\usepackage{xspace} %

\graphicspath{{./media}}

\definecolor{bestcolor}{HTML}{C0E2CA}
\definecolor{sbestcolor}{HTML}{E2EDB9}

\DeclareMathOperator{\atantwo}{atan2}

\newcommand{\ours}{Splat-LOAM\xspace}

\renewcommand{\paragraph}[1]{\vspace{1pt}\noindent{\bf #1}}

\DeclareAcronym{RPE}{
    short = RPE,
    long = Relative Pose Error
}
\DeclareAcronym{SLAM}{
    short = SLAM,
    long = Simultaneous Localization And Mapping
}
\DeclareAcronym{TSDF}{
    short = TSDF,
    long = Truncated Signed Distance Function
}
\DeclareAcronym{SDF}{
    short = SDF,
    long = Signed Distance Function
}
\DeclareAcronym{NVS}{
    short = NVS,
    long = Novel View Synthesis
}
\DeclareAcronym{MLP}{
    short = MLP,
    long = Multi-Layer Perceptron
}
\DeclareAcronym{lc}{
    short = LC,
    long = Loop Closure
}

\newcommand{\bv}{\mathbf{v}}

\newcommand{\bt}{\mathbf{t}}
\newcommand{\bM}{\mathbf{M}}

\newcommand{\bD}{\mathbf{D}}

\newcommand{\bK}{\mathbf{K}}

\newcommand{\bH}{\mathbf{H}}
\newcommand{\bN}{\mathbf{N}}

\newcommand{\bO}{\mathbf{O}}

\newcommand{\bR}{\mathbf{R}}
\newcommand{\bS}{\mathbf{S}}

\newcommand{\bT}{\mathbf{T}}

\newcommand{\cG}{\mathcal{G}}

\newcommand{\cL}{\mathcal{L}}

\newcommand{\depth}{d}

\newcommand{\bbR}{\mathbb{R}}

\newcommand{\bb}{\mathbf{b}}

\newcommand{\bq}{\mathbf{q}}
\newcommand{\bP}{\mathbf{P}}
\newcommand{\bu}{\mathbf{u}}

\newcommand{\bs}{\mathbf{s}}
\newcommand{\bx}{\mathbf{x}}

\newcommand{\bn}{\mathbf{n}}
\newcommand{\bh}{\mathbf{h}}

\newcommand{\bp}{\mathbf{p}}

\newcommand{\bmu}{\boldsymbol{\mu}}
\newcommand{\bnu}{\boldsymbol{\nu}}
\newcommand{\bSigma}{\mathbf{\Sigma}}

\newcommand{\bnabla}{\boldsymbol{\nabla}}
\newcommand{\brho}{\boldsymbol{\rho}}

\def\g2o{$g^2o$}
\def\t2v{\mathrm{log}}
\def\v2t{\mathrm{exp}}
\def\ev2t{\mathrm{ev2t}}

\def\eqref#1{Eq.~(\ref{#1})}

\def\ie{{i.e.}}

\def\etal{\emph{et al.}}
\def\lidar{LiDAR}
\def\lidars{LiDARs}

\newacronym{slam}{SLAM}{Simultaneous Localization and Mapping}
\newacronym{sfm}{SfM}{Structure from Motion}
\newacronym{ba}{BA}{Bundle Adjustment}
\newacronym{sdf}{SDF}{Signed Distance Function}
\newacronym{lo}{LO}{LiDAR odometry}
\newacronym{vbr}{VBR}{A Vision Benchmark in Rome}
\newacronym{nc}{NC}{Newer College Dataset}
\newacronym{ate}{ATE}{Absolute Trajectory Error}
\newacronym{rpe}{RPE}{Relative Pose Error}

%% file: sec/0_abstract.tex
\begin{abstract}
LiDARs provide accurate geometric measurements, making them valuable for ego-motion estimation and reconstruction tasks.
Although its success, managing an accurate and lightweight representation of the environment still poses challenges.
Both classic and NeRF-based solutions have to trade off accuracy over memory and processing times.
In this work, we build on recent advancements in Gaussian Splatting methods to develop a novel \lidar~odometry and mapping pipeline that exclusively relies on Gaussian primitives for its scene representation.
Leveraging spherical projection, we drive the refinement of the primitives uniquely from \lidar~measurements.
Experiments show that our approach matches the current registration performance, while achieving SOTA results for mapping tasks with minimal GPU requirements. This efficiency makes it a strong candidate for further exploration and potential adoption in real-time robotics estimation tasks. Open source will be released at \url{https://github.com/rvp-group/Splat-LOAM}.

\end{abstract}

%% file: sec/1_intro.tex
\vspace{-0.45cm}
\section{Introduction}
\label{sec:intro}
\lidar~sensors provide accurate spatial measurements of the environment, making them valuable for ego-motion estimation and reconstruction tasks. 
Since the measurements already capture the 3d structure, many \lidar~\ac{SLAM} pipelines do not explicitly refine the underlying point representation~\cite{besl1992icp, segal2009gicp, borrmann2008slam6d, ferrari2024mad, dellenbach2021cticp, di2022md}. Moreover, a global map can be obtained by directly stacking the measurements together. However, this typically leads to extremely large point clouds that cannot be explicitly used for online applications. Several approaches attempted to use surfels~\cite{behley2018rss, quenzel2021mars} and meshes~\cite{ruan2023slamesh}. Although these approaches manage to simultaneously estimate the sensor's ego-motion while optimizing the map representation, they result in a trade-off between accuracy, memory usage, and runtime.
\begin{figure}[htbp]
    \centering
    \includegraphics{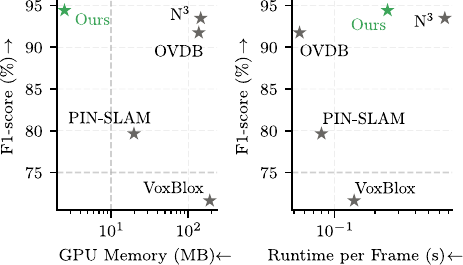}
    \caption{\textbf{Performance overview of \ours.} F1 score to Active Memory [Mb] and Runtime [s]. The plots provide a quantitative comparison between state-of-the-art mapping pipelines, while PIN-SLAM and ours also perform online odometry.}
    \label{fig:motiv}
\end{figure}

The recent advent of NeRF~\cite{mildenhall2020nerf} sparked fresh interest in \ac{NVS} tasks. Specifically, given a set of input views and triangulated points (\ie~obtainable via COLMAP~\cite{schoenberger2016sfm, schoenberger2016mvs}), NeRF learns a continuous volumetric scene function. 
Color and density information is propagated throughout the radiance field by ray-casting pixels from the view cameras and, through a \ac{MLP}, the method learns a representation that ensures multi-view consistency.
Concurrently, NeRFs inspired the computer vision community to tackle the problem of dense visual \ac{SLAM} through \emph{implicit} methods. The pioneering work in this context was iMAP~\cite{sucar2021iccv}, which stores the global appearance and geometry of the scene through a single \ac{MLP}. Despite its success in driving a new research direction, the method suffered from issues related to the limited capacity of the model, leading to low reconstruction quality and catastrophic forgetting during the exploration of larger areas. 
These issues were later handled by shifting the paradigm and by moving some of the appearance and geometric information over a hierarchical feature grid~\cite{zhu2022cvpr} or neural point clouds~\cite{sandstrom2023iccv, liso2024loopy}.
Similarly, these techniques were employed on \lidar~measurements to provide more accurate and lighter explicit dense representations~\cite{zhong2023icra, song2024n3, deng2023nerfloam, loner2023, pan2024tro}.
However, these methods still require tailored sampling techniques to estimate the underlying \ac{SDF} accurately. This bottleneck still poses problems for online execution.

A recent, explicit alternative to NeRFs is 3D Gaussian Splatting (3DGS)~\cite{kerbl3Dgaussians}.
This approach leverages 3D Gaussian-shaped primitives and a differentiable, tile-based rasterizer to generate an appearance-accurate representation. Furthermore, having no need to model empty areas and no neural components, 3DGS earned a remarkable result in accuracy and training speeds. 
Additionally, several approaches further enhanced the reconstruction capabilities of this representation~\cite{Guedon2024, Dai2024GaussianSurfels, Huang2DGS2024}. 
Being fast and accurate, this representation is now sparking interest in dense visual \ac{SLAM}. Recently, 3D Gaussians were employed in several works, yielding superior results over implicit solutions~\cite{Matsuki2024gsslam, yugay2023gaussianslam, zhu2025_loopsplat}.

One issue with Gaussian Splatting relates to the primitive initialization. In areas where few or no points are provided by SFM, adaptive densification tends to fail, often yielding under-reconstructed regions.
\lidar~sensor is quite handy at solving this problem as it provides explicit spatial measurements that can be used to initialize the local representation~\cite{hong2024liv, wu2024mmgaussians}.

However, to our knowledge, no attempt has been made to evaluate the performance of this representation for pure \lidar~data. This technique could prove interesting for visual \ac{NVS} initialization and \lidar~mapping as it could produce a lightweight, dense, and consistent representation.
These insights led us to the development of \ours, the first \lidar~Odometry and Mapping pipeline that only leverages Gaussian primitives as its surface representation.
Our system demonstrates results on par with other SOTA pipelines at a fraction of the computational demands, proving as an additional research direction for real-time perception in autonomous systems.

%% file: sec/2_related_work.tex
\section{Related Work}
\label{sec:related_work}
\paragraph{Classic \lidar{} Odometry and Mapping.} 
\emph{Feature}-based methods that leverage specific points or groups of points to perform incremental registration. For instance,~\cite{zhang2014loam, legoloam2018} track feature points on sharp edges and planar surface patches, enabling high-frequency odometry estimation.
On the other hand, \emph{Direct} methods leverage the whole cloud to perform registration. Specifically, these methods can be categorized based on the subjects of the alignment. \emph{Scan-to-Scan} methods matches subsequent clouds, either explicitly~\cite{besl1992icp, segal2009gicp, borrmann2008slam6d} or through neural 
methods~\cite{Li2019LONetDR, wang2021pwclonet}, while \emph{Scan-to-Model} methods match clouds with either a local or a global map. Typically, the map is represented using points~\cite{ferrari2024mad, dellenbach2021cticp}, surfels~\cite{behley2018rss, quenzel2021mars}, meshes~\cite{ruan2023slamesh}.
Another explored solution project the measurements 
onto a spherical projection plane to leverage visual techniques for ego-motion estimation~\cite{nicolai2016deeplf, zheng2021elo, di2022md, di2023photometric}.

\paragraph{Implicit Methods.}
Concerning mapping only methods, Zhong~\etal~\cite{zhong2023icra} proposed the first method for \lidar~implicit mapping that, given a set of point clouds and the corresponding sensor poses, used hierarchical feature grids to estimate the \ac{SDF} of the scene. Their results demonstrated once again the advantages of such representation for surface reconstruction in terms of accuracy and memory footprint. A similar approach from Song~\etal~\cite{song2024n3} improves the mapping accuracy by introducing \ac{SDF} normal guided sampling and a hierarchical, voxel-guided sampling strategy for local optimization.
Building on these advancements, several \lidar~odometry and mapping techniques were proposed.
Deng~\etal~\cite{deng2023nerfloam} presented the first implicit \lidar~LOAM system using an octree-based feature representation to encode the scene's \ac{SDF}, used both for tracking and mapping. Similarly, Isaacson~\etal~\cite{loner2023} proposes a hierarchical feature grid to store the \ac{SDF} information while using point-to-plane ICP to register new clouds. 
Pan~\etal~\cite{pan2024tro} leverages a neural point cloud representation to ensure a globally consistent estimate.
The new clouds are registered using a correspondence-free point-to-implicit model approach. These methods prove that implicit representation can offer SOTA results in accuracy at the cost of potentially high execution times or memory requirements.
Although targeting a different problem setting, related are also a series of visual neural SLAM methods with RGB or RGBD input~\cite{zhu2022cvpr,sandstrom2023iccv,liso2024loopy,zhang2024glorie,keetha2024splatam,matsuki2024gaussian,yugay2023gaussianslam,sandstrom2024splat}, see \cite{tosi2024nerfs} for a survey.

\paragraph{Gaussian Splatting.} 
Few works tackle the use of \lidar~ within the context of Gaussian Splatting. Wu~\etal~\cite{wu2024mmgaussians} propose a multi-modal fusion system for \ac{SLAM}. Specifically, by knowing the \lidar~to camera rigid pose, the initial pose estimate is obtained through point cloud registration and further refined via photometric error minimization. In this framework, the \lidar~points are leveraged to initialize the new 3D Gaussians.
In a related approach, Hong~\etal~\cite{hong2024liv} proposes a \lidar-Inertial-Visual \ac{SLAM} system. The initial estimate is computed through a \lidar-Inertial odometry. Points are partitioned using size-adaptive voxels to initialize 3D Gaussians using per-voxel covariances. Both the primitives and the trajectory are further refined via photometric error minimization.
\begin{figure*}[htbp]
    \centering
    \includegraphics[width=\textwidth]{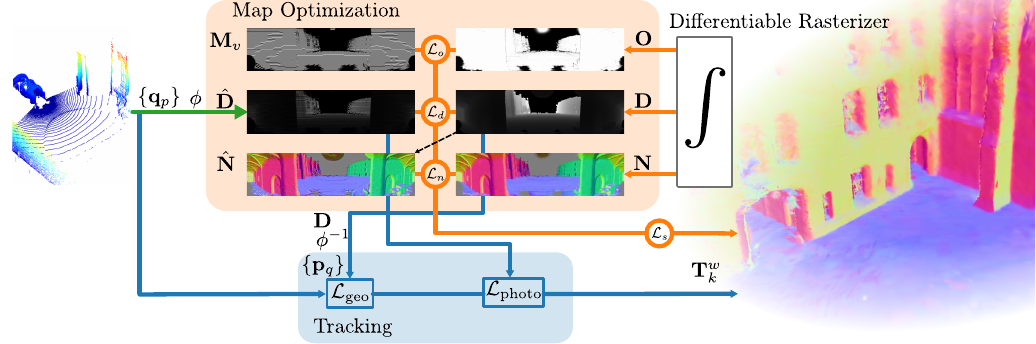}
    \caption{\textbf{\ours Overview.} Given a \lidar~point cloud, we leverage the spherical projection to generate an image-like representation. Moreover, using an ad-hoc differentiable rasterizer, we guide the optimization for structural parameters of 2D Gaussians. The underlying representation is concurrently used to incrementally register new measurements.}
    \label{fig:overview}
\end{figure*}
While these methods introduce \lidar~measurement, 3D Gaussians are still inherently processed by cameras. Recently, Chen~\etal\cite{chen2024lidargs} applies 3D Gaussians for the task of \lidar~\ac{NVS} for re-simulation. The authors propose the use of Periodic Vibrating 3D Gaussian primitives to account for dynamic objects present in the scene. The primitives are initialized using a lightweight \ac{MLP}, and the rasterization is carried out in a spherical frame by computing a per-primitive plane orthogonal to the ray that connects the primitive's mean to the \lidar~frame, thus removing any distortion in the projection process.
Focusing on a different formulation, Jiang~\etal~\cite{jiang2025gslidar} propose a method of \lidar~\ac{NVS} that leverages Periodic Vibrating 2D Gaussian primitives. The primitives are initialized by randomly sampling points and, further optimized using the losses described in~\cite{Huang2DGS2024}, along with a Chamfer loss is introduced to constrain the 3D structures of the synthesized point clouds, and an additional ray-drop term to account for phenomena like non-returning laser pulses. This term is further refined through a U-Net that considers other factors, such as the distance of the surface from the sensor.
Compared to this work, we provide a thorough methodology to render 2D Gaussians on spherical frames while accounting for coordinates singularity and a cloud registration technique to allow for simultaneous odometry and mapping.
To our knowledge, \ours is the first pure \lidar~Odometry and Mapping pipeline that leverages Gaussian primitives both for mapping and tracking.
In sum, our contributions are
\begin{itemize}
\item A differentiable, tile-based rasterizer for 2D Gaussians for spherical frames.
\item A mapping pipeline that allows the merging of sequential \lidar~measurements into a 2D Gaussian representation.
\item A tracking schema that leverages both 3D and 2D representations to register new measurements and estimate the sensor ego-motion.
\end{itemize}

%% file: sec/3_method.tex
\section{Method}
\label{sec:gsloam}
This section introduces our novel LiDAR odometry and mapping method based on 2D Gaussian primitives.
We detail a mapping strategy for initializing, refining, and integrating these primitives alongside a registration method that leverages geometric and photometric cues from the continuous local model for ego-motion estimation. 
Additional details can be found in the supplementary material.

\subsection{Spherical Projection Model}
\label{sec:spherical_projection}
While original Gaussian Splatting leverages pinhole-camera projection to render or refine 3D primitives, LiDAR input provides 360$^\circ$ panoramic input.
To this end, we employ spherical projection to encode \lidar~measurements into an image-like representation that we can use to guide the Gaussian primitives optimization.
A projection is a mapping $\phi : \bbR^3 \rightarrow \Gamma \subset \bbR^2$ from a world
point $\bp = ( x, y, z )^T$ to image coordinates
$\bu = (u, v)^T$.  Knowing the \textit{range} $d=\lVert\bp\rVert$ of an image point $\bu$, we can calculate the inverse mapping $\phi^{-1} : \Gamma \times \bbR \rightarrow \bbR^3$, more explicitly $\bp = \phi^{-1}(\bu, \depth)$. We refer to this
operation as back-projection. To ease the clarity of this work, it is worth mentioning that, compared to the classical pinhole camera, the optical reference frame is rearranged; the $x$-axis points forward, $y$-axis points to the left, and $z$-axis points upwards.
Let $\bK$ be a camera intrinsics matrix that can be computed directly from the point cloud (see Supplementary Material), with function $\psi$ mapping a 3D point to azimuth and elevation. Thus, the spherical projection of a point is given by
\begin{align}
	\phi(\bp) &= \bK \psi(\bp), \label{eq:spherical-projection}\\
	\psi(\bv) &= 
	\begin{bmatrix}
	\atantwo(v_y, v_x) \\
	\atantwo\left(v_z, \sqrt{v_x^2 + v_y^2}\right)\\ 1
	\end{bmatrix}.
\end{align}
We used spherical projection to obtain a range map $\hat\bD$ of the \lidar~point cloud $\left\{\bq_p\right\}_{p=1}^Q$ of size Q.

\subsection{2D Gaussian Splatting}
\label{sec:2dgs}
Our method employs 2D Gaussians as the only scene representation, unifying the required design for accurate, efficient odometry estimation, mapping, and rendering. Due to the inherent thin structure and the explicit encoding of surface normals, 2D Gaussians have demonstrated excellent performance in surface reconstruction \cite{Huang2DGS2024, Dai2024GaussianSurfels}, making them our preferred choice for primitives.

We define a 2D Gaussian by its opacity $o\in\left[0, 1\right]$, centroid $\bmu \in \mathbb{R}^3$, two tangential vectors $\bt_\alpha \in \mathbb{R}^3$ and $\bt_\beta \in \mathbb{R}^3$ defining its plane, and scaling vector $\bs = \left(s_\alpha, s_\beta\right) \in \mathbb{R}^2$ that controls the variance of the Gaussian kernel along the plane~\cite{Huang2DGS2024}. Let $\bt_n = \bt_\alpha\times\bt_\beta$ be the normal of the plane, then the rotation matrix $\bR=\left(\bt_\alpha, \bt_\beta, \bt_n\right) \in \mathbb{SO}(3)$. By arranging the scale factors into a diagonal matrix $\bS \in \mathbb{R}^{3\times3}$, whose last entry is zero, we obtain the following homogeneous transform that maps points $(\alpha, \beta)$ from the splat-space to the world-space $w$:
\begin{align}
    \label{eq:world_T_splat}
    &\bH = \begin{pmatrix}
        s_\alpha\bt_\alpha & s_\beta\bt_\beta & \mathbf{0} & \bmu\\
        0 & 0 & 0 & 1
    \end{pmatrix} = \begin{pmatrix}
        \bR\bS & \bmu \\
        0 & 1
    \end{pmatrix}
\end{align}
Additionally, the Gaussian kernel can be evaluated in the splat-space using:
\begin{equation}
    \mathcal{G}(\alpha,\beta) = \exp\left(-\frac{\alpha^2 + \beta^2}{2}\right).
\end{equation}

\subsubsection{Rasterization}
\label{sec:rast}
We render the 2D Gaussians via $\alpha$-blending as in~\cite{kerbl3Dgaussians}.
This technique requires, for each pixel, the list of primitives to blend, sorted by range. Instead of computing the per-pixel sorting of primitives, which is too expensive, we subdivide the image into a grid of  $16\times 16$ tiles and, concurrently, generate a per-tile list of primitives sorted from closer to further. Using the method described in \cref{sec:bbcomp}, we compute the bounding box for each primitive and generate a per-tile list of primitives to be rasterized.
Then, for each tile, we sort the $T$ Gaussians based on their range, and finally, for each pixel $\bu$, we integrate them using $\alpha$-blending from front to back to obtain a range $d$, normal $\bn$ and opacity $o$ values, as follows:
\begin{align}
    \label{eq:ab_depth}
    &d=\sum_{i=1}^{T}o_i\mathcal{G}_id_i\prod_{j=1}^{i-1}\left(1-o_j\mathcal{G}_j\right)\\
    \label{eq:ab_normal}    &\bn=\sum_{i=1}^{T}o_i\mathcal{G}_i\bt_{n_i}\prod_{j=1}^{i-1}\left(1-o_j\mathcal{G}_j\right)\\
    \label{eq:ab_opacity}
    &o=\sum_{i=1}^{T}o_i\mathcal{G}_i\prod_{j=1}^{i-1}\left(1-o_j\mathcal{G}_j\right)
\end{align}
We do not rely on the local affine approximation proposed in~\cite{kerbl3Dgaussians} due to numerical instabilities for slanted views and distortions for larger primitives. Instead, we rely on an explicit ray-splat intersection proposed in~\cite{Weyrich2007}. This is efficiently found by locating the intersection of three non-parallel planes.\\
\subsubsection{Ray-splat Intersection} 
\label{sec:raysplat-intersection}
Let $\bv = \phi^{-1}\left(\bK^{-1}\bu\right)$ be the normalized direction of pixel $\bu$. We parametrize $\bv$ as the intersection of two orthogonal planes:
\begin{align}
    &\mathbf{h}_x = \frac{\mathbf{v}\times\mathbf{u}_z}{\lVert\mathbf{v}\times\mathbf{u}_z\rVert}  \enspace, & \mathbf{h}_y = \mathbf{h}_x\times\mathbf{v}  \enspace,
\end{align}
where $\mathbf{u}_z$ is the unit $z$-direction vector. 
Then, we represent the planes in the splat's space as:
\begin{align}
    \label{eq:planes_to_splat}
    &\bh_\alpha = \left(\bT^c_w \bH\right)^T\bh_x  \enspace, & \bh_\beta = \left(\bT^c_w \bH\right)^T\bh_y \enspace,
\end{align}
where $\bT^k_w \in \mathbb{SE}(3)$ describes the \textit{world} in the $k$-th sensor reference frame.
We express the intersection of the two planes bundle as:
\begin{equation}
    \bh_\alpha\left(\alpha,\beta,1,1\right)^T = \bh_\beta\left(\alpha,\beta,1,1\right)^T = 0 \enspace.
\end{equation}
The solution $\alpha$ and $\beta$ is then computed by solving the homogeneous linear system:
{\small
\begin{align}
    &\alpha(\bu_s) = \frac{\bh_\alpha^2\bh_\beta^4 - \bh_\alpha^4\bh_\beta^2}
    {\bh_\alpha^1\bh_\beta^2 - \bh_\alpha^2\bh_\beta^1}  \enspace,
    &\beta(\bu_s) = \frac{\bh_\alpha^4\bh_\beta^1 - \bh_\alpha^1\bh_\beta^4}
    {\bh_\alpha^1\bh_\beta^2 - \bh_\alpha^2\bh_\beta^1}  \enspace.
\end{align}}

\subsubsection{Bounding Box Computation}
\label{sec:bbcomp}
\begin{figure}[t]
    \centering
    \includegraphics[width=\columnwidth]{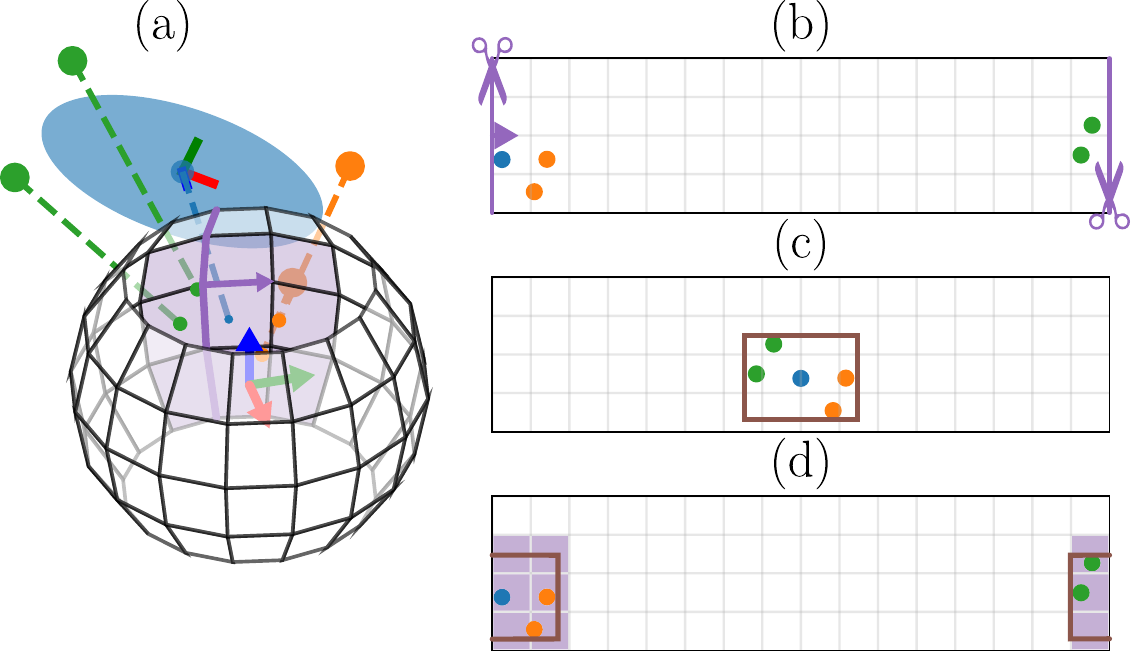}
    \caption{\textbf{Bounding box computation for near-singularity splats.}~\textbf{(a)} shows the 3D configuration of a splat that approximately lies behind the camera. \textbf{(b)} shows the corresponding spherical image with the projected bounding box vertices. The distortion is removed by shifting the vertices along the horizontal direction to align the projected center to the image center. \textbf{(c)} being far from the coordinate singularity, we compute the maximum extent of the splat. \textbf{(d)} we revert the shift and propagate to the corresponding tiles via an offset from the central vertex, matching with the tiles highlighted on (a).}
    \label{fig:aabb}
\end{figure}
To leverage the efficient tile-based rasterizer, we need to know the tiles that should rasterize each primitive. Typically, this is achieved by estimating a bounding box around the projected central point of the splat. On spherical images, however, we need to account for the coordinate singularity at the horizontal boundaries $\{0, W\}$. \Cref{fig:aabb} describes the approach we designed to compute the splat image-space bounding box for spherical projection models. First, we compute the $3\sigma$ splat-space bounding box and use \cref{eq:spherical-projection} to obtain its image-space vertices.
Moreover, we shift the vertices to match the central vertex to the image center to ensure that no vertex is projected near the coordinate singularity at the horizontal image boundary. 
Eventually, we compute the horizontal extent of the splat, revert the previous rotation, and propagate the splat's ID to the nearby tiles while accounting for the coordinate singularity. This allows us to obtain consistent bounding boxes even when the splat lies approximately behind the sensor's frame.\\

\subsection{Odometry And Mapping}
\label{sec:odometry-and-mapping}

Following the modern literature of RGB-D SLAM \cite{zhu2025_loopsplat, sandstrom2024splat, liso2024loopy, yugay2023gaussianslam}, we rely on keyframing to optimize local maps. We choose this approach for the following advantages: first, continuous integration over the same model can have adverse effects on artifacts and, more importantly, on runtime. Instead of decreasing the local density, we reset to a new local model if certain conditions are met, while also restricting the number of frames joining the optimization stage to allow effective online processing
Based on this, we define each local map as an individual Gaussian model $\bP^s$:
\begin{equation}
    \bP^s = \left\{\cG(\bmu, \bSigma, o)|i=1,\ldots,N\right\}.
\end{equation}
\subsubsection{Local model initialization}
\label{sec:local_model_init}

We initialize a new local model using the input LiDAR point cloud whenever necessary, such as at system startup or when visibility conditions require it. As a first step, we generate a valid pixel mask via the indicator function $\mathbbm{1}[\cdot]$ as 
\begin{equation}
    \bM_v= \mathbbm{1}[\hat{\bD}>0 ],
\end{equation} 
and compute the range gradients $\bnabla_{\bD}$ to construct a weight map over the range image. We use a weighted sampling of $n_d$ points to prioritize complex regions.
Splat positions are computed by back-projecting the range image, while their orientations are initialized by directing surface normals toward the sensor center to enhance initial visibility. 

\subsubsection{Local Model Refinement}
We perform a limited number of refinement iterations $n_o$ on the most recent keyframes. Unlike~\cite{zhu2025_loopsplat, yugay2023gaussianslam}, we employ a geometric distribution-based sampling scheme that guarantees at least $40\%$ probability of selecting the most recent keyframe and progressively decreases the likelihood of choosing the older ones.
To filter out artifacts caused by ray-drop phenomena in the \lidar~measurements, we only consider valid pixels to refine the local model parameters $\bx$.
We start by applying a densification strategy similar to the one described in \cref{sec:local_model_init}, with two extra terms. The first one, $\bM_n = \mathbbm{1}[\bO_i \leq \lambda_{d,o} ]$, target newly discovered areas while the second, $\bM_e = \mathbbm{1}[\lvert \bD_i - \hat{\bD}_i\rvert\geq\lambda_{d,e} ]$, target under-reconstructed regions.

To optimize the geometric consistency of the local model, we employ a loss term that minimizes the $L_1$ error:
\begin{equation}
    \label{eq:loss_range}
    \cL_d = \sum_{\bu \in \bM_v}\rho_d \rVert \bD(\bu,\bx) - \hat \bD(\bu) \lVert,
\end{equation}
where $\rho_d$ is a weight function dependent on the measurement's range.
In addition, we employ a self-regularization term to align the splat's normals to the surface normals $\bN$ estimated by the gradients of the range map $\bD$~\cite{Huang2DGS2024}:
\begin{equation}
    \label{eq:loss_normal}
    \cL_n = \sum_{\bu \in \bM_v} 1 - \mathbf{n}^T\mathbf{N}\big(\bD(\bu, \bx)\big)
\end{equation}
Furthermore, to promote the expansion of splats over uniform surfaces, we introduce an additional term that operates on the opacity channel of the rasterized images.
Specifically, we drive the splats to cover the areas of the image containing valid measurements by correlating the opacity image $\bO$ with the valid mask $\bM_v$.
\begin{equation}
    \label{eq:loss_opacity}
    \cL_o = \sum_{\bu \in \bM_v} -\log\big(\bO(\bu, \bx)\big).
\end{equation}
While $\cL_o$ allows the splats to grow, it can also cause some splats to become extremely large in unobserved areas.
We employ an additional regularization term on the scaling parameter of all primitives $N$ to compensate for this effect. We propose a novel regularization that allows the splat to extend up to a certain value $\tau_s$ before penalizing its growth:
{\small
\begin{equation}
    \cL_s = \left\{
    \begin{array}{cl}
         \tau_s - \max\left(s_\alpha, s_\beta \right) & \text{if}~\max\left(s_\alpha ,s_\beta \right) > \tau_s \\
         0 & \text{otherwise} 
    \end{array}\right .
\end{equation}}
We found this term to be more effective rather than directly minimizing the deviation from the average \cite{zhu2025_loopsplat, yugay2023gaussianslam, Matsuki2024gsslam} as it allows for anisotropic splats that are particularly useful for mapping details and edges, and provides more control on the density and structure of the model. 

The final mapping objective function is defined as:
\begin{equation}
    \label{eq:loss_full}
    \mathcal{L}_\mathrm{map} = \mathcal{L}_d + \lambda_o \mathcal{L}_o + \lambda_n\mathcal{L}_n + \lambda_s \sum^N_{i=1} \mathcal{L}_{s_i} \enspace,
\end{equation}
with $\lambda_o, \lambda_n, \lambda_s \in \bbR$ being loss weights.
We do not perform an opacity reset step, which could lead to catastrophic forgetting.

\subsubsection{Frame-To-Model Registration}
\label{sec:frame-to-model-registration}
Each time a new keyframe is selected, we sample the local model by back-projecting the rasterized range image $\bD$ onto a point cloud $\{ \bp_q \}_{q=1}^{W \times H}$ at its estimated pose.
We design an ad-hoc tracking schema to benefit from both geometric and photometric consistencies provided by the LiDAR and the rendered local model. Hence, the total odometry loss is composed of the sum of both residuals:
\begin{equation}
\cL_\mathrm{odom} = \cL_\mathrm{geo} + \cL_\mathrm{photo}.
\end{equation}

\paragraph{Geometric Registration.} 
To associate geometric entities, we employ a PCA-based kd-tree~\cite{ferrari2024mad}. The kd-tree is built on the back-projected rendered range map $\{ \bp_q \}_{q=1}^{W \times H}$ and segmented into tree leaves, corresponding to planar patches. Each leaf corresponds to $l = \left < \bp_l, \bn_l \right >$, that is, the mean point $\bp_l \in \bbR^3$ and the surface normal $\bn_l \in \bbR^3$. The geometric loss $\cL_\mathrm{geo}$ represents the sum of the point-to-plane distance between the mean leaf $\bp_l$ and the point of the current measurement point cloud $k$ with $\{ \bq_p \}_{p=1}^{Q}$, along the normal $\bn_l$ expressed in the local reference frame $\bT_w^k$. Specifically, we have:
\begin{equation}
\cL_\mathrm{geo} = \sum_{p, q \in \{a\}} \rho_\mathrm{Huber} \left((\bT_w^k \bn_{l_q})^T ( \bT_w^k \bq_p - \bp_{q_i}) \right),
\end{equation}
where $w$ is the global frame, $k$ is the local frame, $\{a\}$ is the set of leaf associations with the point from the measurement point cloud, and $\rho_\mathrm{Huber}$ is the Huber robust loss function.\\

\paragraph{Photometric Registration.} Leveraging the rendered range map $\bD$, we employ photometric registration for subpixel refinement, minimizing the photometric distance between the rendered and the spherical projected query point cloud $\hat \bD$. The photometric loss is formulated as:
{\small\begin{equation}
\cL_\mathrm{photo} = \sum_{\bu} \Big\lVert \rho_\mathrm{Huber} \Big(\bD(\bu) - \hat \bD \underbrace{\big( \phi \big( \bT_w^k \phi^{-1}( \bu, \hat d ) \big) \big)}_{\bu'} \Big) \Big\rVert^2.
\end{equation}}
The evaluation point $\bu'$ in the query image $\hat \bD$ is computed by first back-projecting the pixel $\bu$, applying the transform $\bT_w^k$, and then projecting it back. 
Pose updates $\delta$ are parameterized as local updates in the Lie algebra $\mathfrak{se}(3)$. Therefore, the transformation $\bT_k^w$ of the local reference frame, expressed in the global reference frame, is updated as:
\begin{equation}
\bT_k^w \leftarrow \bT_k^w \exp(\delta).
\end{equation}
This update is carried out using a second-order Gauss-Newton method. Local updates ensure that rotation updates are well-defined~\cite{engels2006bundle}.

%% file: sec/4_experiments.tex
\section{Experiments}
\label{sec:experiments}
\begin{table*}[!htbp]
    \centering
    \footnotesize
    \setlength{\tabcolsep}{3.2pt}
    \begin{tabular}{@{}lcccccccccccccccc@{}}
        \toprule
         \multirow{2}{*}{Dataset}&\multicolumn{4}{c}{Newer College\cite{zhang2021multicamera}} & \multicolumn{12}{c}{Oxford Spires\cite{tao2024oxspires}}\\
         \cmidrule(lr){2-5}\cmidrule(lr){6-17}
         &\multicolumn{4}{c}{quad-easy}&\multicolumn{4}{c}{keble-college02}&\multicolumn{4}{c}{bodleian-library-02}&\multicolumn{4}{c}{observatory-01}\\
         \cmidrule(lr){2-5}\cmidrule(lr){6-9}\cmidrule(lr){10-13}\cmidrule(lr){14-17}
         Approach & Acc$\downarrow$ & Com$\downarrow$ & C-l1$\downarrow$ & F-score$\uparrow$
         & Acc$\downarrow$ & Com$\downarrow$ & C-l1$\downarrow$ & F-score$\uparrow$
         & Acc$\downarrow$ & Com$\downarrow$ & C-l1$\downarrow$ & F-score$\uparrow$
         & Acc$\downarrow$ & Com$\downarrow$ & C-l1$\downarrow$ & F-score$\uparrow$\\
         \midrule
         OpenVDB\cite{museth2013vdb} & 11.45 & \cellcolor{sbestcolor}4.38 & \cellcolor{sbestcolor}7.92 & 88.85 & 7.46 & \cellcolor{bestcolor}\textbf{6.92} & \cellcolor{sbestcolor}7.19 & 91.74 & \cellcolor{sbestcolor}10.34 & \cellcolor{sbestcolor}4.68 & \cellcolor{bestcolor}\textbf{7.51} & 89.68 & 9.58 & \cellcolor{bestcolor}\textbf{9.60} & \cellcolor{sbestcolor}9.59 & \cellcolor{sbestcolor}86.16\\
         VoxBlox\cite{oleynikova2017voxblox}&20.36&12.64&16.5&64.63&15.81&14.25&15.03&71.63&18.92&11.56&15.24&58.77&15.09&15.15&15.12&70.45\\
         $N^3$-Mapping\cite{song2024n3}&\cellcolor{bestcolor}\textbf{6.32}&9.75&8.04&\cellcolor{sbestcolor}94.54&\cellcolor{sbestcolor}6.21&\cellcolor{sbestcolor}7.82&\cellcolor{bestcolor}\textbf{7.01}&\cellcolor{sbestcolor}93.47&\cellcolor{bestcolor}\textbf{10.16}&5.62&7.89&\cellcolor{bestcolor}\textbf{90.36}&\cellcolor{bestcolor}\textbf{8.27}&\cellcolor{sbestcolor}10.44&\cellcolor{bestcolor}\textbf{9.35}&\cellcolor{bestcolor}\textbf{87.94}\\
         PIN-SLAM\cite{pan2024tro}&15.28&10.5&12.89&88.05&13.73&9.94&11.83&79.65&14.34&7.14&10.74&82.71&16.91&12.07&14.49&72.31\\
         Ours&\cellcolor{sbestcolor}6.64&\cellcolor{bestcolor}\textbf{4.09}&\cellcolor{bestcolor}\textbf{5.37}&\cellcolor{bestcolor}\textbf{96.74}&\cellcolor{bestcolor}\textbf{6.18}&8.69&7.43&\cellcolor{bestcolor}\textbf{94.41}&10.87&\cellcolor{bestcolor}\textbf{4.33}&\cellcolor{sbestcolor}7.6&\cellcolor{sbestcolor}90.09&\cellcolor{sbestcolor}9.35&11.76&10.56&83.04\\
         \bottomrule
    \end{tabular}
    \caption{\textbf{Reconstruction quality evaluation.} The pipelines were run with ground-truth poses. Voxel size is set to $20$~cm and F-score is computed with a $20$~cm error threshold. \ours yields competitive mapping performance on both the Newer College\cite{zhang2021multicamera} and Oxford Spires\cite{tao2024oxspires} datasets and outperforms most competitive approaches. }
    \label{tab:exp_mapping}
\end{table*}
In this section, we report the results of our method on different publicly available datasets. We evaluate our pipeline on both tracking and mapping.
We recall that, to our knowledge, this is the first Gaussian Splatting \lidar{} Odometry and Mapping pipeline, and no direct competitors are available. 
We compared our approach with other well-known geometric and neural implicit methods.
The experiments were executed on a PC with an Intel Core i9-14900K @ 3.20Ghz, 64GB of RAM, and an NVIDIA RTX 4090 GPU with 24 GB of VRAM.

For the odometry experiments, we evaluate over several baselines. The first one is a basic point-to-plane ICP odometry, as a minimal example. Then, we include two geometric pipelines: 
SLAMesh simultaneously estimates a mesh representation of the scene via Gaussian Processes and perform registration onto it~\cite{ruan2023slamesh}. Moreover, MAD-ICP leverages a forest of PCA-based KD-Trees to perform accurate registration. 
Furthermore, we include PIN-SLAM as SOTA baseline for implicit \lidar~\ac{SLAM}~\cite{pan2024tro}. It leverages neural points as primitive and interleaves an incremental learning of the model's \ac{SDF} and a correspondence-free, point-to-implicit registration schema.
We also highlight that we could not run the official implementation of NeRF-LOAM~\cite{deng2023nerfloam}, and thus, we do not include it in the evaluation.

We evaluate the mapping capabilities of our method against popular mapping SOTA pipelines.
We include OpenVDB~\cite{museth2013vdb} and VoxBlox~\cite{oleynikova2017voxblox} as ``classic'' baselines. OpenVDB provides a robust volumetric data structure to handle 3D Points, while VoxBlox combines adaptive weights and grouped ray-casting for an efficient and accurate \ac{TSDF} integration. 
Additionally, we include two neural-implicit mapping pipelines. $N^3$-Mapping is a neural-implicit non-projective \ac{SDF} mapping pipeline~\cite{song2024n3}. It leverages normal guidance to produce more accurate \ac{SDF}s, leading to SOTA results for offline \lidar~mapping. 
PIN-SLAM \cite{pan2024tro} is included also for the mapping experiments. In fact, using marching cubes, it can produce a mesh from the underlying implicit \ac{SDF}. Below, we report the datasets and the evaluation metrics employed.
We highlight that we could not run the official implementation of SLAMesh~\cite{ruan2023slamesh} with ground-truth poses. Hence, we do not include the pipeline for quantitative comparisons.

\subsection{Datasets}
\begin{figure}[b]
    \centering
    \includegraphics[width=\linewidth]{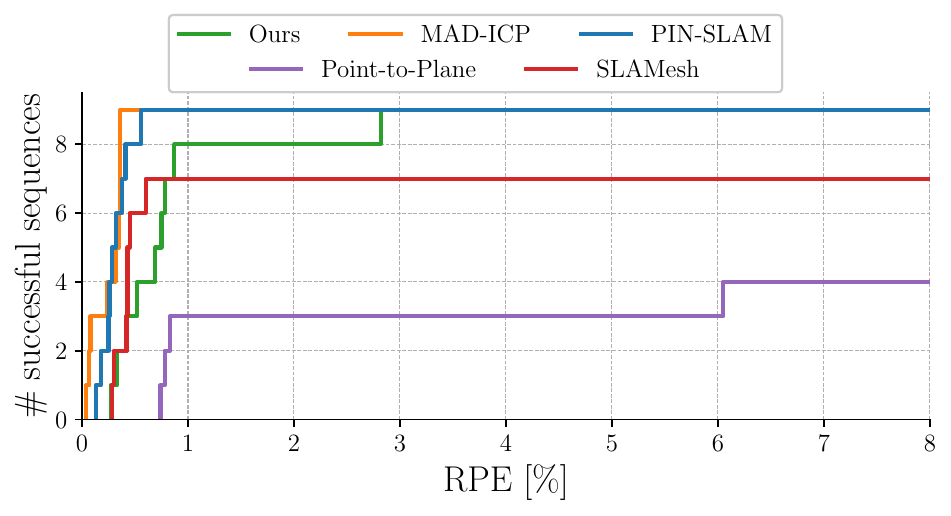}
    \caption{\textbf{RPE evaluation.} Number of successful sequences across RPE thresholds. It includes the sequences of Newer College~\cite{zhang2014loam}, VBR~\cite{bg2024vbr}, Oxford Spires~\cite{tao2024oxspires} and Mai City~\cite{vizzo2021icra}.}
    \label{fig:rpe}
\end{figure}

\noindent
We used the following four publicly available datasets:
\begin{itemize}
\setlength\itemsep{4pt}
\item \textbf{\gls{nc} \cite{zhang2021multicamera}:} Collected with a handheld Ouster OS0-128 \lidar{} in structured and vegetated areas. Ground truth was generated using the Leica BLK360 scanner, achieving centimeter-level accuracy over poses and points in the map.
\item \textbf{\gls{vbr} \cite{bg2024vbr}:} Recorded in Rome using OS1-64 (car-mounted) and OS0-128 (handheld) \lidars{}, capturing large-scale urban scenarios with narrow streets and dynamic objects. Ground truth trajectories were obtained by fusing \lidar{}, IMU, and RTK GNSS data, ensuring centimeter-level accuracy over the poses.
\item \textbf{Oxford Spires \cite{tao2024oxspires}:} Recorded with a hand-held \emph{Hesai QT64} LiDAR featuring a $360^\circ$ horizontal FoV, $104^\circ$ vertical FoV, $64$ vertical channels, and $60$ meters of range detection. Similar to \cite{zhang2014loam}, each sequence includes a prior map obtained via a survey-grade 3D imaging laser scanner, used for ground-truth trajectory estimation and mapping evaluation. Specifically, we choose the \emph{Keble College}, \emph{Bodleian Library}, and \emph{Radcliffe Observatory} sequences to include both indoor and outdoor scenarios with different levels of detail.
\item \textbf{Mai City \cite{vizzo2021icra}:} A synthetic dataset captured using a car-like simulated \lidar{} with $120$ meters of range detection. The measurements are generated via ray-casting on an underlying mesh, providing error-free, motion-undistorted data. We select the \emph{01} and \emph{02} sequences, which capture similar scenarios with different vertical resolutions.
\end{itemize}
\subsection{Evaluation}
\label{sec:eval}
We use \ac{RPE} computed with progressively increasing delta steps to evaluate the odometry accuracy. Specifically, we adapt the deltas to the trajectory length to provide a more meaningful result~\cite{bg2024vbr}. 
Differently, to evaluate mapping quality, we use several metrics~\cite{mescheder2019cvpr}: Accuracy (Acc) measures the mean distance of points on the estimated mesh with their nearest neighbors on the reference cloud. Completeness (Comp) measures the opposite distance, and Chamfer-l1 (C-l1) describes the mean of the two. Additionally, we use the F-score computed with $20$ cm error threshold.
\subsection{Ablation Study}
\label{sec:ablation}
Our approach employs Gaussian primitives for LiDAR odometry estimation and mapping, yielding results comparable to state-of-the-art methods while significantly enhancing computational efficiency.
In this section, we evaluate some key aspects of our pipeline and evaluate their contributions.\\

\begin{figure}[b]
    \centering
    \includegraphics[width=1.0\columnwidth]{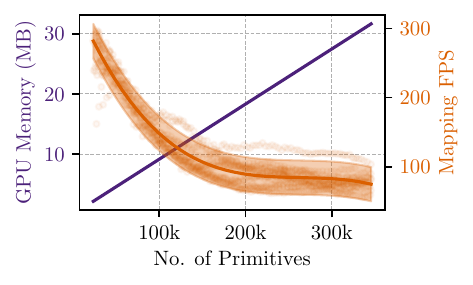}
    \caption{\textbf{Memory and Runtime Analysis.} The plot relates the used GPU Memory and the mapping iteration frequency with the number of active primitives. The measurements were obtained over the longest sequence we reported: campus~\cite{bg2024vbr}.}
    \label{fig:ablation_runtime}
\end{figure}

\paragraph{Memory and Runtime Analysis.}
In \cref{fig:ablation_runtime}, we report how the increment of active primitives affects the active GPU memory requirements and the per-iteration mapping frequency. It shows an experiment ran over the large-scale \emph{campus} sequence~\cite{bg2024vbr} where we set a maximum of $100$ keyframes per local map and sample, at most, $50\%$ of points for the incoming point cloud.
It's possible to notice that the mapping FPS remains stable between $200$k and $300$k primitives. This is most likely related to the saturation of the rasterizer.
\begin{figure*}[htbp]
    \centering
    \includegraphics[width=\linewidth]{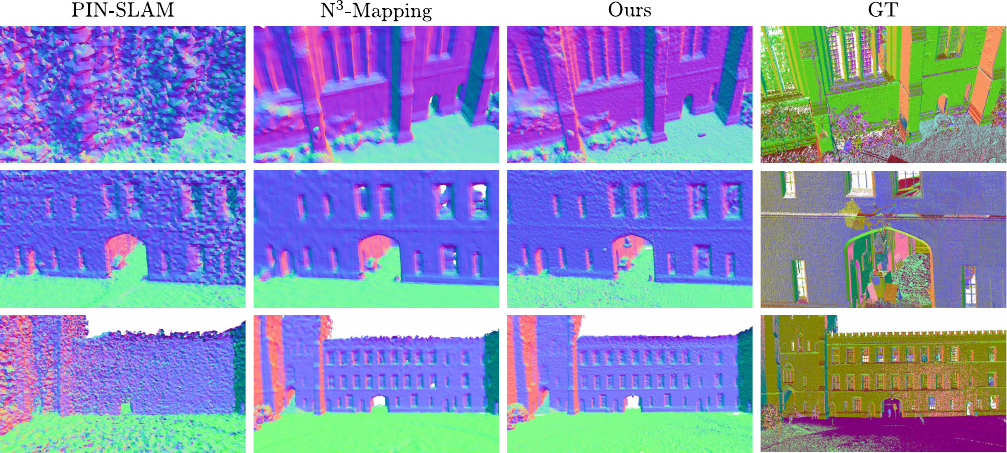}
    \caption{\textbf{Comparison of Mesh Reconstruction.} The figure shows reconstruction results for \textit{quad-easy} sequence from the newer college dataset. Our method recovers a geometry with much higher data fidelity. PIN-SLAM lacks many details and exhibits a large level of noise. $N^3$-Mapping performs more similar to ours, but oversmoothes fine geometric details.}
    \label{fig:mapping_mesh_comparison}
\end{figure*}
\\

\begin{figure}
    \centering
    \includegraphics[width=\linewidth]{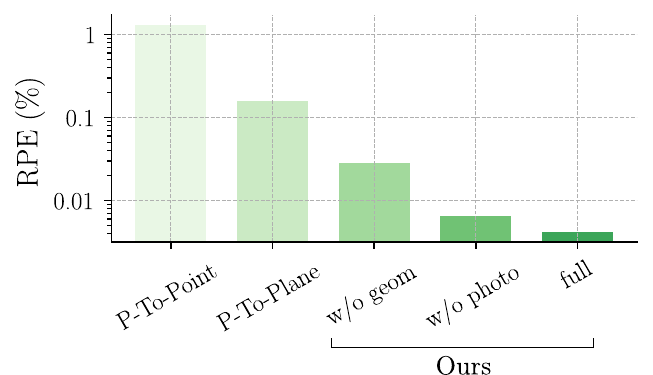}
    \caption{\textbf{Ablation on registration methods}. The plot reports the RPE (\%) of several tested registration methods on the quad-easy sequence. Enabling both geometric and photometric factors in sequence, provides a more robust estimate. the quad-easy sequence~\cite{zhang2021multicamera}}
    \label{fig:ablation_tracking}
\end{figure}

\paragraph{Odometry.}
\cref{fig:ablation_tracking} shows the results for different tracking methods over our scene representation.
Using both geometric and photometric components, we achieve a better result than using point-to-point or point-to-plane. The last three bars show an ablation of geometric and photometric loss components. The results are best for the joint use of both terms which support our design choice.

\paragraph{Mesh generation.}
To generate a mesh from the underlying representation, we sample $n_m$ points from each keyframe rendered range and normal maps. Moreover, similar to~\cite{Guedon2024}, we accumulate the points into a single point cloud and generate the mesh using Poisson Reconstruction~\cite{Kazhdan2006}. 
\cref{tab:ablation_mesh} shows the results of two additional methods for meshing that we tested: the Poisson Reconstruction of the Gaussian primitives' centers and normals and the \ac{TSDF} integration using the rendered depth and normal maps from the keyframes.
\begin{table}[t]
    \centering
    \footnotesize
    \setlength{\tabcolsep}{8.6pt}
    \renewcommand{\arraystretch}{1.0}
    \begin{tabular}{lcccc}
    \toprule
    Extraction Method&Acc$\downarrow$&Com$\downarrow$&C-l1$\downarrow$&F-score$\uparrow$\\
    \midrule
    Marching Cubes\cite{museth2013vdb}&16.76&\cellcolor{sbestcolor}5.53&11.14&76.76\\
    Poisson (centers)&\cellcolor{sbestcolor}10.15&6.70&\cellcolor{sbestcolor}8.43&\cellcolor{sbestcolor}92.33\\
    Ours&\cellcolor{bestcolor}\textbf{6.64}&\cellcolor{bestcolor}\textbf{4.09}&\cellcolor{bestcolor}\textbf{5.37}&\cellcolor{bestcolor}\textbf{96.74}\\
    \bottomrule
    \end{tabular}
    \caption{\textbf{Ablation on Meshing Methods.} We report mapping results with varying meshing methods on the quad-easy sequence~\cite{zhang2021multicamera}. Our method yields consistently better results.}
    \label{tab:ablation_mesh}
\end{table}

%% file: sec/5_conclusions.tex
\vspace{-0.2cm}
\section{Conclusion}
\label{sec:conclusion}
We present the first \lidar~Odometry and Mapping pipeline that leverages 2D Gaussian primitives as the sole scene representation. Through an ad-hoc tile-based Gaussian rasterizer for spherical images, we leverage \lidar~measurements to optimize the local model. Furthermore, we demonstrate the effectiveness of combining a geometric and photometric tracker to register new \lidar~point clouds over the Gaussian local model. The experiments show that our pipeline obtains tracking and mapping scores that are on par with the current SOTA at a fraction of the computational demands.\\
\paragraph{Future Work.}
We plan on improving \ours by simultaneously estimating the sensor pose and velocity to compensate for motion skewing of the \lidar~measurements. Moreover, we plan on including \ac{lc} to improve the pose estimates, along with the mapping accuracy.

%% file: sec/6_supp.tex
\clearpage
\setcounter{page}{1}
\setcounter{section}{0}
\counterwithin{figure}{section} %
\counterwithin{table}{section}
\renewcommand{\thesection}{\Alph{section}} %
\maketitlesupplementary
\noindent In this supplementary material, we provide additional details on various components and design choices that were not fully elaborated in the main paper. These include the computation of the camera matrix, the rationale behind the bounding box computation, the incorporation of color features, and the analytical derivation of our spherical rasterizer.
\vspace{0.3cm}
\section{Camera Matrix}
Using hard-coded field of views from the sensor's datasheet may lead to empty areas inside the image (\ie, when parts of the FoV contain no observable environment or, due to rounding errors). To solve these issues, we recompute the field of views, along with a camera matrix, for each input \lidar~point cloud.

Let $\{\bp_q\}_{q=1}^{N}$ be a set of points expressed at the sensor origin. Let $\{\bv_q = \left(\gamma, \theta, 1\right)^T=\psi(\bp_q)\}_{q=1}^{N}$ be the same set of points expressed in spherical coordinates. From this representation, we can directly estimate the camera matrix $\bK$ as follows. First, we compute the maximum horizontal and vertical angular values within the set:
\begin{align}
    &\gamma_m = \min_\gamma \left\{\bv_q\right\} & \gamma_M = \max_\gamma \left\{\bv_q\right\},\\
    &\theta_m = \min_\theta \left\{\bv_q\right\} & \theta_M = \max_\theta \left\{\bv_q\right\}.
\end{align}
Moreover, we compute the horizontal $\mathrm{FoV}_h = \gamma_M - \gamma_m$ and vertical $\mathrm{FoV}_h = \theta_M - \theta_m$ field of views and, provided an image size $\left(H, W\right)$, we estimate the camera matrix as:
\begin{equation}
    \bK\left(\left\{\bp_q\right\}\right) = 
    {
    \footnotesize
    \begin{pmatrix}
        -\frac{W-1}{\mathrm{FoV}_h} & 0 & \frac{W}{2}\left(1 + \frac{\gamma_M + \gamma_m}{\mathrm{FoV}_h}\right)\\
        0 & -\frac{H-1}{\mathrm{FoV}_v} & \frac{H}{2}\left(1 + \frac{\theta_M + \theta_m}{\mathrm{FoV}_v}\right)\\
        0 & 0 & 1
    \end{pmatrix}
    }
\end{equation}
\vspace{0.3cm}
\section{Bounding Box}
In this section, we report a supplementary study concerning the computation of the tightly aligned bounding box on spherical images.
Efficiently computing the tightly aligned bounding box for a splat on the view space requires solving a $4$-th-order polynomial due to the complexity of the underlying manifold. While fixing the azimuth angle $\gamma$ results in a planar surface in $\mathbb{R}^3$, fixing the altitude angle $\theta$ leads to a cone subspace in $\mathbb{R}^3$.
To find the tightly aligned bounding box, we should search the spherical coordinates $\left(\gamma,\theta\right)$ that exactly intersect tangentially the splat space at a distance $3\sigma$ from the origin. Projecting the $\alpha$-plane onto the splat frame results in a line, and the intersection condition can be solved via a linear equation. Projecting the $\gamma$-cone onto the splat's frame results in a parametric 2D conic equation. Enforcing two tangent solutions leads to a polynomial of $4$th-degree. Given the small image sizes of LiDAR images and the relatively high cost of solving higher-order polynomials, we opt for an easier but less optimal solution.
We relax the tight constraint and obtain an image-space bounding box by projecting the individual bounding box vertices.
This typically results in a bounding box that includes more pixels but is faster in computation.
\vspace{0.3cm}
\section{On color features}
Modern LiDARs provide a multitude of information on the beam returns. Specifically, they provide details concerning the mean IR light level (ambient) and the returned intensity (intensity). Through this information, it is also possible to compute the \emph{reflectivity} of the surface using the inverse square law for Lambertian objects in far fields. 
Throughout this study, we opted to omit the color information to focus on the geometric reconstruction capabilities of our approach. Moreover, we think incorporating intensity and reflectivity channels can pose a challenge due to the inherent nature of LiDARs. Both properties cannot be explicitly related to a portion of the space but rather from a combination of the surface and sensor position with respect to the former.
\vspace{0.3cm}
\section{Rasterizer Details}
In this section, we describe the process of rasterization over spherical images.
First, we provide a detailed analysis of the rasterization process for a Gaussian primitive.
Furthermore, we provide the analytical derivatives for the components of the process.
Recall that a Gaussian primitive $\mathcal{G}$ is defined by its centroid $\bmu\in\mathbb{R}^3$, its covariance matrix decomposed as a rotation matrix $\bR\in\mathbb{SO}(3)$ and a scaling matrix $\bS$, and its opacity $o\in\mathbb{R}$.
To obtain the homogeneous transform that maps points $\left(\alpha,\beta\right)$ from the splat-space to the sensor-space $c$, we decouple the axes of the rotation matrix $\bR = \left(\bt_\alpha,\bt_\beta, \bt_n\right)$ and the per-axis scaling parameters $\bs=\left(s_\alpha,s_\beta\right)$, and assume $\bT_w^c\in\mathbb{SE}(3)$ be the transform the world in camera frame. By concatenating $\bT_w^c$ with \cref{eq:world_T_splat}, we obtain the following transform:
\begin{equation}
    \bT_{4\times3} = \bT_w^c\bH = 
    \begin{pmatrix}
        \bb_\alpha & \bb_\beta & \bb_c\\
        0&0&1
    \end{pmatrix},
\end{equation}
where $\bb_c = \bR_w^c\bmu + \bt_w^c$. We omit the third column of $\bT$, which is zeroed by construction.
\subsection{Forward Process}
In this section, we describe the rasterization process for a pixel $\bu = \left(u, v\right)$. We assume that primitives are already pre-sorted.
As described in \cref{sec:raysplat-intersection}, we compute the orthogonal planes in splat-space by pre-multiplying each plane by $\bT$:
\begin{align}
    &\bh_\alpha = \bT^T\bh_x &\bh_\beta = \bT^T\bh_y,
\end{align}
and compute the intersection point $\hat\bp$:
\begin{align}
    &\hat\bp = \bh_\alpha\times\bh_\beta\\
    &\hat\bs = \left(\hat s_\alpha, \hat s_\beta\right) = \left(\frac{\hat\bp_x}{\hat\bp_z}, \frac{\hat\bp_y}{\hat\bp_z}\right)^T.
\end{align}
We use $\hat\bs$ to estimate two quantities. First, we measure the Gaussian kernel at the intersection point $\mathcal{G}\left(\hat\bs\right)$ to compute the Gaussian density
\begin{equation}
    \alpha = o\mathcal{G}\left(\hat\bs\right),
\end{equation}
and second, we compute the range as
\begin{align}
    &\bnu = \hat s_\alpha \bb_\alpha + s_\beta \bb_\beta + \bb_c\\
    &d = \lVert\bnu\rVert.
\end{align}
We follow \cref{eq:ab_depth}, \cref{eq:ab_normal}, and \cref{eq:ab_opacity} to $\alpha$-blend the sorted Gaussians and compute the pixel contributions.
\vspace{0.3cm}
\subsection{Gradient Computation}
From the rasterizer perspective, we assume to already have the per-pixel channel derivatives, namely the depth $\frac{\partial\mathcal{L}}{\partial d}\in\mathbb{R}$ and normal $\frac{\partial\mathcal{L}}{\partial\bn}\in\mathbb{R}^3$.
To improve the readability, each partial derivative also includes its dimension using the $\frac{\partial A}{\partial B}\rvert_{\text{dim}(A)\times \text{dim}(B)}$ notation.
Finally, we show the computation for the $k$-th Gaussian over the $m$ Gaussians contributing to the pixel.
\\
First, we derive the gradients w.r.t the density:
\begin{align}
    &\frac{\partial d}{\partial d_k}\Bigg|_{1\times 1} = d_kA_k - \frac{B_{d,k}}{1 - \alpha_k}\\
    &\frac{\partial \bn}{\partial \bn_k}\Bigg|_{1\times 3} = \bn_kA_k - \frac{B_{\bn, k}}{1 - \alpha_k}\\
    &\frac{\partial\mathcal{L}}{\partial \alpha_k}\Bigg|_{1\times 1} = \frac{\partial\mathcal{L}_d}{\partial d}\frac{\partial d}{\partial \alpha_k} + \frac{\partial\mathcal{L}_n}{\partial \bn}\frac{\partial \bn}{\partial \bn_k},
\end{align}
where $A_k=\prod_{i=1}^{k-1}(1-\alpha_i)$, $B_{d,k}=\sum_{i>k}d_i\alpha_iA_i$, and $B_{\bn,k}=\sum_{i>k}\bn_i\alpha_iA_i$. We leverage the sorting of the primitives to efficiently compute these values during the back-propagation of the gradients.
\\
Furthermore, we propagate the gradients to the homogeneous transform matrix $\bT$ from the intersection of planes $\hat\bp_k$:
\begin{align}
    \begin{split}
    \frac{\partial\mathcal{L}}{\partial \hat\bs_k}\Bigg|_{1\times 2} 
    &= \frac{\partial\mathcal{L}}{\partial\alpha}\frac{\partial \alpha}{\partial \hat\bs_k} 
    + \frac{\partial\mathcal{L}}{\partial d_k}\frac{\partial d_k}{\partial \hat\bs_k}\\
    &= -\frac{\partial\mathcal{L}}{\partial \alpha}\alpha_k\hat\bs_k^T
    + \frac{\partial\mathcal{L}}{\partial d_k}\frac{\alpha_kA_k}{d_k}
    \begin{pmatrix}
        \bnu^T\bb_\alpha\\
        \bnu^T\bb_\beta
    \end{pmatrix}^T
    \end{split}\\
    \begin{split}
        \frac{\partial \mathcal{L}}{\partial \hat\bp_k}\Bigg|_{1\times 3} 
        &= \frac{\partial \mathcal{L}}{\partial \hat\bs_k}\frac{\partial \hat\bs_k}{\partial\hat\bp_k}\\
        &= \frac{1}{\hat\bp_z}
        \begin{pmatrix}
        \partial\mathcal{L}/\partial \hat\bs_\alpha
        \\
        \partial\mathcal{L}/\partial \hat\bs_\beta
        \\
        -\frac{
        \hat\bp_x\partial\mathcal{L}/\partial \hat\bs_\alpha
        +\hat\bp_y\partial\mathcal{L}/\partial \hat\bs_\beta
        }{\hat\bp_z}
        \end{pmatrix}^T.
    \end{split}
\end{align}
Thus, we can derive the gradients over the matrix $\bT$. We keep the three accumulators $\frac{\partial\mathcal{L}}{\partial\bb_\alpha}$, $\frac{\partial\mathcal{L}}{\partial\bb_\beta}$, and $\frac{\partial\mathcal{L}}{\partial\bb_c}$ decoupled to correctly integrate the contributions from each pixel:
\begin{equation}
    \brho_\alpha = \left(\frac{\partial\mathcal{L}}{\partial \bp_k}\times\bh_\alpha\right) \quad\quad
    \brho_\beta = \left(\frac{\partial\mathcal{L}}{\partial \bp_k}\times\bh_\beta\right)\\
\end{equation}
\begin{align}
    \begin{split}
        \frac{\partial\mathcal{L}}{\partial \bb_\alpha}\Bigg|_{1\times 3} 
        &= \frac{\partial\mathcal{L}}{\partial\hat\bp_k}\frac{\partial\hat\bp_k}{\partial\bb_\alpha}
        + \frac{\partial\mathcal{L}}{\partial d_k}\frac{\partial d_k}{\partial\bb_\alpha}\\
        &= -\brho_{\beta, 1} \bh_x + \brho_{\alpha, 1} \bh_y + \frac{\mathcal{L}}{\partial d_k}\frac{\hat s_\alpha}{d_k}\bnu^T
    \end{split}
    \\
    \begin{split}
        \frac{\partial\mathcal{L}}{\partial \bb_\beta}\Bigg|_{1\times 3} 
        &= \frac{\partial\mathcal{L}}{\partial\hat\bp_k}\frac{\partial\hat\bp_k}{\partial\bb_\beta}
        + \frac{\partial\mathcal{L}}{\partial d_k}\frac{\partial d_k}{\partial\bb_\beta}\\
        &= -\brho_{\beta, 2} \bh_x + \brho_{\alpha, 2} \bh_y + \frac{\mathcal{L}}{\partial d_k}\frac{\hat s_\beta}{d_k}\bnu^T
    \end{split}
    \\
    \begin{split}
        \frac{\partial\mathcal{L}}{\partial \bb_c}\Bigg|_{1\times 3} 
        &= \frac{\partial\mathcal{L}}{\partial\hat\bp_k}\frac{\partial\hat\bp_k}{\partial\bb_c}
        + \frac{\partial\mathcal{L}}{\partial d_k}\frac{\partial d_k}{\partial\bb_c}\\
        &= -\brho_{\beta, 3} \bh_x + \brho_{\alpha, 3} \bh_y + \frac{\mathcal{L}}{\partial d_k}\frac{1}{d_k}\bnu^T,
    \end{split}
\end{align}
where $\brho_{k, i}$ is the $i$-th element of $\brho_k$.\\
Finally, we compute the gradients w.r.t. the Gaussian parameters.
\begin{align}
    \frac{\partial\mathcal{L}}{\partial\bR_k}\Bigg|_{3\times 3} &= 
    \begin{pmatrix}
        s_\alpha\frac{\partial\mathcal{L}}{\partial \bb_\alpha}^T &
        s_\beta\frac{\partial\mathcal{L}}{\partial \bb_\beta}^T &
        \frac{\partial\mathcal{L}}{\partial \bn_k}^T
    \end{pmatrix}
    \bR_w^c
    \\
    \frac{\partial\mathcal{L}}{\partial\bS_k}\Bigg|_{1\times 2} &=
    \begin{pmatrix}
        \frac{\partial\mathcal{L}}{\partial\bb_\alpha}\bR_w^c\bR_{\left[1\right]}
        &\frac{\partial\mathcal{L}}{\partial\bb_\alpha}\bR_w^c\bR_{\left[1\right]}
    \end{pmatrix}
    \\
    \frac{\partial\mathcal{L}}{\partial\bmu_k}\Bigg|_{1\times 3} &=
    \frac{\partial\mathcal{L}}{\partial\bb_c}\bR_w^c,
    \\
    \frac{\partial\mathcal{L}}{\partial o_k}\Bigg|_{1\times 1} &=
    \frac{\partial\mathcal{L}}{\partial \alpha_k}\exp\left(-\frac{1}{2}\bs_k^T\bs_k\right)
\end{align}
where $\bR_{\left[i\right]}$ is the $i$-th column of $\bR$.
\begin{figure*}[htbp]
    \centering
    \includegraphics[width=\linewidth]{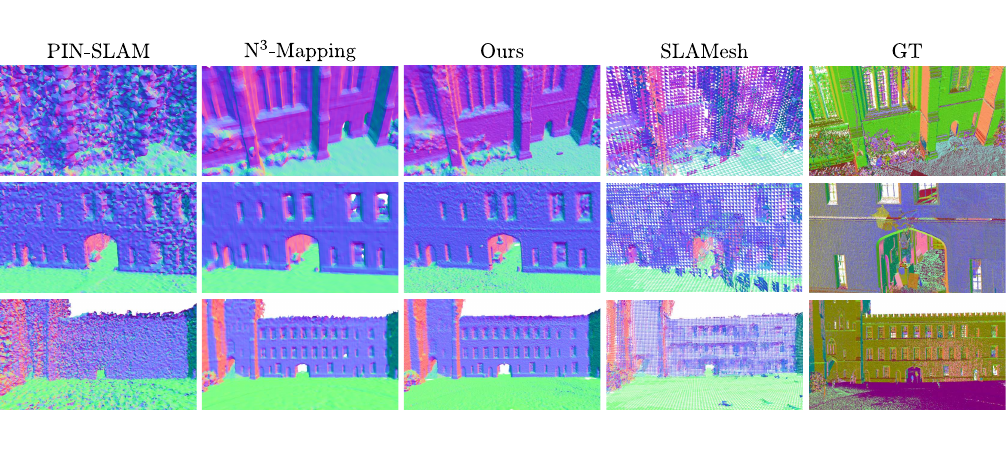}
    \caption{\textbf{Qualitative Mapping Results.} The images show the mapping results for different pipelines in the quad-easy sequence. We also include the SLAMesh pipeline, which was evaluated on a self-estimated trajectory.}
    \label{fig:slamesh}
\end{figure*}
\vspace{0.3cm}
\section{Additional Qualitative Comparison}
\begin{figure*}[htbp]
    \centering
    \includegraphics[width=0.85\linewidth]{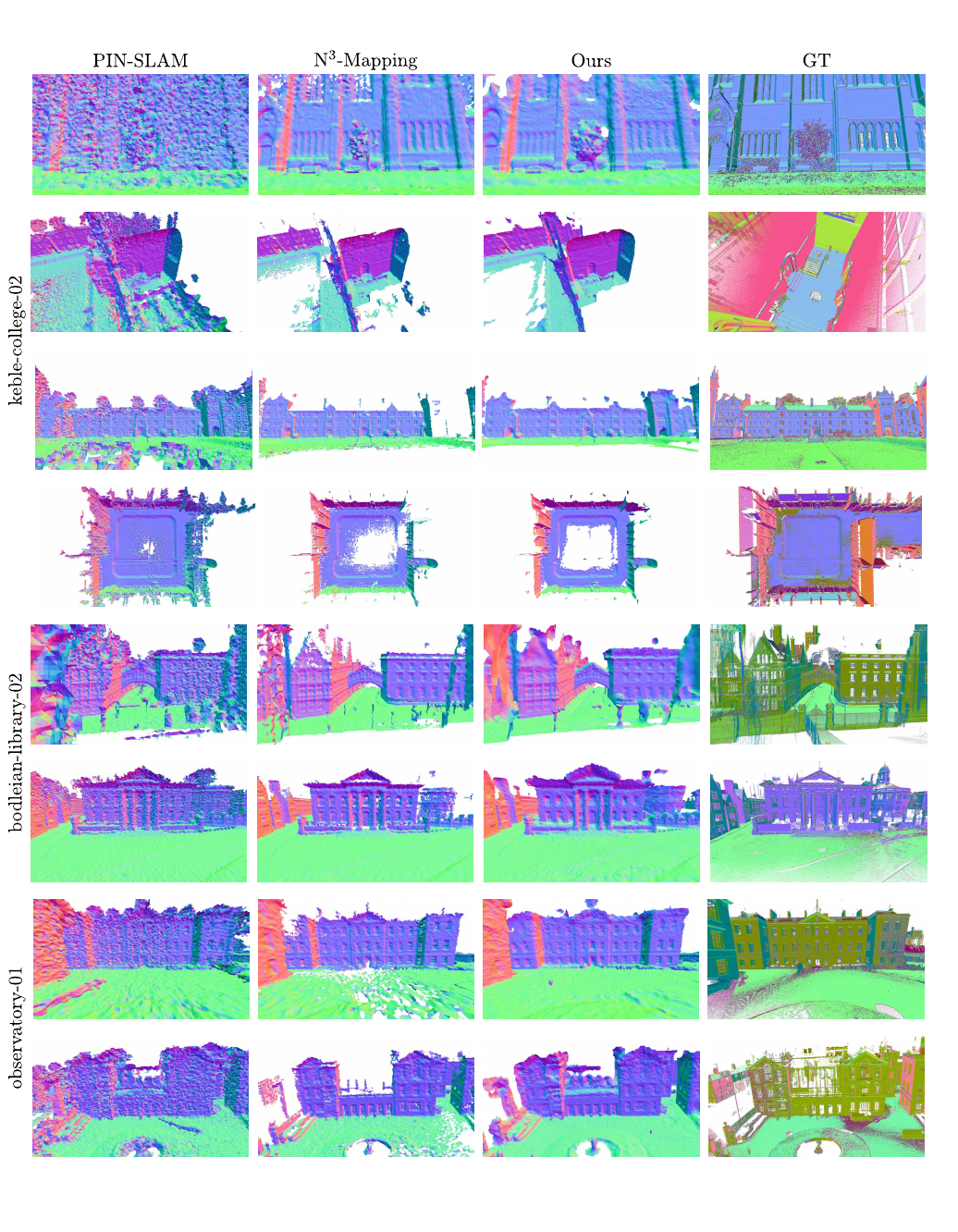}
    \caption{\textbf{Qualitative Mapping Results.} The images show the mapping results for different pipelines in the Oxford Spires dataset \cite{tao2024oxspires}.}
    \label{fig:ox-spires}
\end{figure*}

In this section, we show additional results over the meshing reconstruction. \Cref{fig:slamesh} includes the results we obtained using the work of Ruan~\etal~\cite{ruan2023slamesh}. We remark that we did not include these results in the manuscript as we could not run the official implementation over the Ground Truth trajectory.
Additionally, \Cref{fig:ox-spires} shows the reconstruction results over the Oxford Spires dataset~\cite{tao2024oxspires}.
\vspace{0.3cm}
\section{Motion Distortion}
\begin{figure}[htbp]
    \centering
    \includegraphics{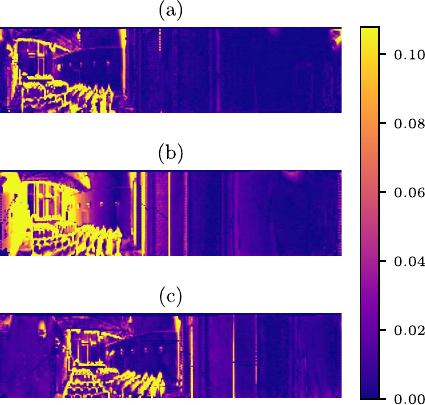}
    \caption{\textbf{Effects of motion distortion during registration.} The images show the projective range error between our model and the incoming measurement \textbf{(a)} before a rotation, \textbf{(b)} during a rotation, and \textbf{(c)} after the rotation. The images are resized over the horizontal axis for visibility purposes. The error is expressed in meters.}
    \label{fig:motion-deskew-error}
\end{figure}

Throughout the experiments, we noticed that \ours is very sensitive to the motion distortion effect caused by the continual acquisition of \lidars. \Cref{fig:motion-deskew-error} shows how the projective error over the estimated model changes while the sensor rotates during the acquisition, hindering both the registration and mapping phases. We plan to compensate for the motion distortion effect by simultaneously estimating the sensor pose and velocity.
